\documentclass[10pt,twocolumn]{article}
\newif\ifreview \reviewfalse
\usepackage[a4paper,top=2.2cm,bottom=2.4cm,left=1.9cm,right=1.9cm,columnsep=0.65cm,headsep=0.7cm,footskip=1cm]{geometry}
\usepackage[T1]{fontenc}
\IfFileExists{XCharter.sty}{\usepackage{XCharter}}{\usepackage{charter}}
\IfFileExists{sourcesanspro.sty}{\usepackage[scale=0.9]{sourcesanspro}}{\usepackage[scaled=0.88]{helvet}}
\usepackage{microtype}
\usepackage{xcolor}
\usepackage{graphicx}
\usepackage{booktabs}
\usepackage{setspace}
\usepackage[super,comma,sort&compress]{natbib}
\usepackage{titlesec}
\usepackage{needspace}
\usepackage{fancyhdr}
\usepackage{caption}
\usepackage{lineno}
\usepackage{longtable,array,calc}
\usepackage{amsmath,amssymb}
\usepackage{newunicodechar}
\newunicodechar{±}{\ensuremath{\pm}}
\newunicodechar{×}{\ensuremath{\times}}
\newunicodechar{−}{\ensuremath{-}}
\newunicodechar{≤}{\ensuremath{\leq}}
\newunicodechar{≥}{\ensuremath{\geq}}
\newunicodechar{Ž}{\v{Z}}
\newunicodechar{é}{\'{e}}
\newunicodechar{ö}{\"{o}}
\newunicodechar{–}{--}
\newunicodechar{—}{---}
\usepackage[hidelinks,pdftitle={System 1 agents: a method for local, reliable and fast agentic AI}]{hyperref}
\ifreview \linenumbers \else  \fi

\titleformat{\section}{\needspace{4\baselineskip}\sffamily\bfseries\small}{}{0pt}{\MakeUppercase}
\titlespacing*{\section}{0pt}{2.4ex plus .5ex minus .2ex}{0.9ex plus .1ex}
\titleformat{\subsection}[runin]{\normalfont\bfseries}{}{0pt}{}[.]
\titlespacing*{\subsection}{0pt}{1.4ex plus .3ex minus .2ex}{0.5em}
\DeclareCaptionLabelSeparator{bar}{\enspace|\enspace}
\fancypagestyle{first}{\fancyhf{}\fancyfoot[R]{\sffamily\scriptsize\thepage}}

\providecommand{\real}[1]{#1}
\begin{document}
\thispagestyle{first}
\twocolumn[{%
\begin{flushleft}
{\sffamily\scriptsize\MakeUppercase{Research Article}\par}
\vspace{0.9em}
{\fontsize{24}{28}\selectfont System 1 agents: a method for local, reliable and fast agentic AI\par}
\vspace{1.1em}
{\sffamily\small\setstretch{1.25}
Eleftherios~Tzanis\textsuperscript{1,2,*},
Michail~E.~Klontzas\textsuperscript{1,2,3}\par}
\end{flushleft}
\vspace{0.6em}
\noindent\rule{\textwidth}{0.6pt}\par\vspace{0.9em}
\noindent\begin{minipage}[t]{0.25\textwidth}{\sffamily\bfseries\small Abstract\par}\end{minipage}%
\begin{minipage}[t]{0.75\textwidth}
\small\setstretch{1.05}\setlength{\parskip}{0.45em}\setlength{\parindent}{0pt}
Agentic artificial intelligence could automate analyses of sensitive
clinical data, but current agents depend on large language models
(LLMs). Cloud LLMs expose patient data. Local LLMs hallucinate, and an
invented value in a tool call propagates through every later step. Here
we present a method to build System 1 agents: local agents that call
tools without generating text. A deterministic compiler turns each
tool's schema and the request into multiple-choice questions whose
options come from the request and the schema, then a one-pass decision
model answers them. On 1,952 requests to 157 imaging and biomedical
tools, the best System 1 agent chose the right tool, or correctly
declined, for 97.0\% of requests, significantly ahead of all 18 local
LLM-agent settings (best 95.2\%). The System 1 agent filled 97.0\% of
6,056 stated argument values. No System 1 call contained invented
content, against 5.5--19.7\% of LLM calls. The System 1 Agents framework is freely available as open-source software at \url{https://github.com/eltzanis/SystemOneAgents}.
\end{minipage}
\par\vspace{1.0em}
\noindent\rule{\textwidth}{0.3pt}\par\vspace{1.6em}
}]
{\let\thefootnote\relax
\footnotetext{\noindent\sffamily\scriptsize\setstretch{1.0}%
\textsuperscript{1}Artificial Intelligence and Translational Imaging (ATI) Lab, Department of Radiology, School of Medicine, University of Crete, Heraklion, Crete, Greece.
\textsuperscript{2}Division of Radiology, Department of Clinical Science Intervention and Technology (CLINTEC), Karolinska Institute, Huddinge, Sweden.
\textsuperscript{3}Department of Medical Imaging, University Hospital of Heraklion, Crete, Greece.
\textsuperscript{*}Corresponding author: Eleftherios Tzanis, PhD. E-mail: eleftherios.tzanis@ki.se; etzanis@uoc.gr; tzaniseleftherios@gmail.com. ORCID: 0000-0003-0353-481X.}}
\section*{Main}

Agentic artificial intelligence (AI) systems plan and execute analyses
by calling software tools\citep{ref1,ref2,ref3}. In clinical research
and practice, such agents could prepare imaging studies, train and apply
models, and query biomedical databases on request. The Model Context
Protocol (MCP) gives tools a common interface for this purpose, where
each tool publishes a name, a description and a JSON schema of its
arguments\citep{ref4}.

Two obstacles keep such agents out of clinical use. The first is
privacy. The most capable language models run as cloud services, so
every request and the data it refers to leave the institution. For
patient data, this conflicts with data protection rules and
institutional policy\citep{ref5,ref6}. Local models remove this
risk, because data never leave the site\citep{ref7}. The second
obstacle is reliability. Local language models are smaller, and an agent
built on any language model writes each tool call token by token. It can
call a tool that does not exist, use an argument the tool lacks or fill
an argument with a value the user never gave\citep{ref8,ref9,ref10,ref11}. In
an analysis pipeline, such a hallucinated value does not fail visibly.
It changes the result of that step, and every later step builds on it.
Reasoning modes reduce some errors but multiply the text generated per
call and, together with it, time and energy\citep{ref12,ref13}.

Nonetheless, most of a tool call is not open-ended and tools are listed
in the catalogue. Almost every value a correct call needs is written in
the request, follows from it by a fixed rule, such as a unit conversion,
or is listed in the tool's schema. What remains is a series of choices.
Specifically, which tool, which written value for which argument,
whether to call at all. Decision models are built for such choices. They
read a state, a question and a finite list of options, and return a
probability for each option in one forward pass\citep{ref14,ref15,ref16,ref17}.
Dual-process accounts of cognition call this kind of fast, single-step
judgement System 1\citep{ref18}.

Here we present a method to build System 1 agents, local agents in which
every decision is a multiple-choice question answered by a decision
model, and no text is generated. We describe the architecture, the
development method and the system we built with it for biomedical
research. We then compare six System 1 agents with nine local LLM-based
agents, in accuracy, reliability and efficiency.
\section*{Results}

\subsection*{A method to build System 1 agents}
The method rests on the principle that the model judges, and code
executes verbatim what is requested. A System 1 agent therefore has
three components with one responsibility each (\textbf{Fig. 1a}): (i)
the agent, (ii) the compiler and (iii) the decision model.

The agent controls the workflow. It asks one question over every tool of
the catalogue plus `none of these'. The System 1 model outputs a
probability score for each option. If the two most probable tools are
within a probability of 0.3, it queries which of the two has parameters
for the values the request gives. It then asks whether the chosen tool
can do what is requested, and declines below a probability of 0.5.
Subsequently, it passes the chosen tool and the request to the compiler.
In case a required value is missing, the agent does not fill it.
Instead, it shows the user a fixed prompt, the argument's name and its
schema description, and waits for the value. When the value is present
it sends the call. The prompt is a template filled from the schema and
no stochastic generation happens at any step.

The compiler is deterministic code that turns the chosen tool's JSON
schema and the request into questions with finite options. An
enumeration is asked over its values, a boolean over `on', `off' or `not
mentioned'. Paths, numbers, names and other free values are asked over
candidates that the compiler extracts from the request and the schema,
plus `not specified'. Candidates include values derived by fixed rules,
such as unit conversions, numbers written as words and paths composed
from written parts. List items are chosen one at a time, so candidates
compete. An optional value is set only when the probability on the
request's values reaches 0.7, and a value equal to the documented
default is not passed. A required value that the request does not give
is never filled and instead the compiler reports it to the agent as
missing.

The decision model answers each question with a probability for every
option, in one forward pass. A question has four parts, a state, a
question text, an option list and a kind (choice or yes/no). Any
decision model that answers this interface can be used unchanged.

This division makes hallucinated content impossible by design. A
decision model can only pick among the options it is shown. Every tool
in a System 1 call therefore comes from the catalogue, every argument
name from the schema and every value from the request, the schema or the
agent's policy. Every decision carries a probability that thresholds
turn into declines and prompts for missing values. On the same hardware
and software, the same input gives the same output, and a run can be
replayed exactly from its stored answers and audited question by
question. \textbf{Figure 1b} follows one request through these steps.

We developed the compiler with a loop that keeps model failures and
compiler failures apart (Methods). A compiler oracle, which replaces the
decision model with a component that knows the label, shows whether the
compiler can express each labelled call. Requests the oracle cannot
complete reveal values that no question offers. These are fixed by
general rules about how requests write values and not by rules for a
particular tool. A change is accepted only if it raises the summed
number of correct calls across models on development requests, loses no
earlier reference request beyond a fixed tolerance and survives
leave-one-out runs.

\subsection*{A local System 1 agent for biomedical research}
We applied the method to multiple biomedical research fields (Methods).
A total of 157 tools of 18 MCP servers for image classification,
segmentation, radiomics, AutoML, a computed tomography foundation model,
proteomics, metabolomics, genomics, single-cell and spatial
transcriptomics, clinical trials and chemistry. We used six published
decision models with 4 to 35 billion parameters, without fine-tuning
(\textbf{Extended Data Table 1}). Each ran locally on one GPU through
llama.cpp.

We compared the System 1 agents with nine open-weight LLM-based agents,
each with reasoning on and off (\textbf{Extended Data Table 2}). Three
of the LLMs are the base models of five of the decision models. Each
LLM-based agent ran as distributed, on the same GPU, with the request as
the only message and every tool as a native function. The evaluation set
comprised 2,126 requests: 2,003 custom prompts and 123 prompts taken
verbatim from published research studies\citep{ref19,ref20}. Together
they contain 1,645 complete requests, which state 6,056 argument values,
307 out-of-scope requests, which no tool can serve, and 174
underspecified requests, which omit a required value. All 123 published
prompts are complete requests. The 1,952 complete and out-of-scope
requests (1,829 custom and 123 published) form the main benchmark for
accuracy; the 174 underspecified requests were evaluated separately. A
scorer tool was developed to judge every agent (Methods).

\subsection*{Comparison by category}
System 1 agents were more accurate in most pairwise comparisons. They
never invented content and, instead of inventing a missing required
value, left it for the user to supply. On declining out-of-scope
requests and on schema validity, the optimal LLM-agent configurations
were comparable. Three of the four LLM-agent configurations that
declined requests as often as System 1 agents, left 13.7-36.5\% of
complete requests without a call. At higher accuracy, OpenJev-based
System 1 agent was faster and used less energy than the best LLM-based
agent configuration. On argument filling, five LLM-based agents were not
significantly different from the best System 1 agent, and the cheapest
LLM-based agent without reasoning, matched the cheapest System 1 agents
on energy and time. \textbf{Table 1} summarizes the comparison between
System 1 agents and LLM-based agents.

\subsection*{Accuracy}
We measured accuracy in three ways on the main benchmark of 1,952
requests (1,645 complete and 307 out-of-scope). Tool calling, counts the requests for which the
agent called the labelled tool or, when the request was out of scope,
declined. Argument filling, counts the 6,056 values that the complete
requests state, with a value considered right only in a call to the
right tool. A request is fully correct when its entire outcome is right.
For a complete request, the agent must call the labelled tool, provide
all stated values correctly, and leave no required arguments empty. For
an out-of-scope request, the agent must decline. Values not specified in
the request are not counted as errors in this assessment, as we
evaluated invented values separately (see Reliability section).

The six System 1 agents reached 91.7-97.0\% accuracy on tool calling,
89.8-97.0\% on argument filling and 81.4-93.9\% on fully correct request
assessment (\textbf{Fig. 2} and \textbf{Extended Data Table 3}).
pplx-decider-27b reached 97.0\%, 97.0\% and 93.9\% (95\% confidence
interval (CI) 92.7-94.8\%), and OpenJev 96.1\%, 96.1\% and 92.9\%,
respectively. The best LLM-based agent, Qwen3.5-27B with reasoning off,
reached 95.1\%, 97.2\% and 91.3\% accuracy on the tool calling, argument
filling and fully correct criteria, respectively. Neither reasoning
setting consistently outperformed the other across LLMs. Enabling
reasoning improved full correctness for five models but reduced it for
four.

We compared each System 1 agent with each LLM-agent in each mode, 108
pairs per measure, with Holm correction (\textbf{Fig. 3}). On fully
correct requests, System 1 was significantly ahead in 73 pairs and the
LLM in 11, with no significant difference in 24. pplx-decider-27b-based
agent was ahead of all 18 LLM-based agents and OpenJev-based agent of
17. All 11 LLM leads were over the two weakest System 1 agents. The
advantage was primarily in tool selection, with System 1 agents
outperforming the LLM-based agents in 76 pairs compared with just 5
pairs in which the LLM-based agents performed better. On argument
filling, the larger LLMs were closer (46 System 1 leads, 19 LLM leads,
43 ties). The two pplx-decider models were ahead of their own base model
run as an agent, by 2.6-9.6 percentage points on fully correct requests;
decider-35b-a3b-based System 1 agent was behind its base model with
reasoning on (\textbf{Extended Data Table 4}). By tool group, System 1
led most on medical imaging tools whose calls carry many technical
values, such as nnU-Net\citep{ref21}, and the LLM-based agents
filled values better for identifier-heavy biomedical databases such as
ClinicalTrials.gov and ChEMBL\citep{ref22} (\textbf{Extended Data
Table 5}).

\subsection*{Reliability}
No System 1 call contained hallucinated content (\textbf{Fig. 4a} and
\textbf{Extended Data Table 7}). Across 10,888 System 1 calls, every
tool and argument name was in the catalogue, and none of 33,945 argument
values was ungrounded. We counted a value as ungrounded when it matched
neither the labelled value, nor text or a number written in the request,
nor an option the compiler offered for that argument, nor the tool's
schema. This test counts any value written anywhere in the request as
grounded, so it undercounts hallucination. The LLM-based agents never
called a nonexistent tool, but three of the nine agents used argument
names absent from the schema, in 19-53 calls per setting. A total of
5.5--19.7\% of LLM-based agent calls contained at least one invented
value. Most invented values were settings the user never stated, such as
hyperparameters, file paths and options. Reasoning did not remove them.

System 1 agents did not guess missing values. On the 174 underspecified
requests, they chose the right tool, filled the stated values and
reported the missing required argument, for which the agent prompts the
user, in 86.2-93.7\% of cases (\textbf{Fig. 4b}). In 2-6 cases per
agent, they filled a missing required value with another value written
in the request, which represents a wrong choice, not an invention. The
LLM-based agents behaved correctly in at most 5.7\% of cases. They
mostly answered without a call or filled the missing value with an
invented one in up to 62.1\% of underspecified requests. Even the most
accurate LLM setting invented the missing value in 27.6\% of them
(\textbf{Table 1}).

System 1 agents declined 87.0-94.5\% of the 307 out-of-scope requests
and left 1.0-3.3\% of complete requests without a call (\textbf{Fig.
4c}). The LLM-based agents declined 41.4-97.4\% and left 0.0-36.5\%
without a call. The two that declined most, left 24.3\% and 36.5\% of
complete requests without a call. Of the three kinds of out-of-scope
request, those for which a tool performs the task but not on the given
input were the hardest for every agent. In 22 of the 10,888 System 1
calls, a value written in the request was placed where the schema
requires another form, such as an identifier pattern. The LLM-based
agents made such calls at a similar rate (\textbf{Table 1}).

\subsection*{Efficiency}
System 1 agents used 0.072-0.407 Wh of GPU energy per fully correct
request, against 0.198-2.481 Wh for LLM-based agents with reasoning and
0.080-0.853 Wh without (\textbf{Fig. 5} and \textbf{Extended Data Table
6}). A request took 1.2-10.1 s with System 1, in comparison to 2.6-26.1
s for LLM-based agents with reasoning and 1.1-9.2 s without. At the top
of the accuracy range, OpenJev-based System 1 agent was fully correct on
92.9\% of requests, with 2.6 s per request and 0.228 Wh per fully
correct request, against 91.3\%, 4.5 s and 0.413 Wh for the best
LLM-based agent setting. A System 1 agent asked
about 12 questions per request and generated no text, whereas the
LLM-based agents generated 123-340 tokens per request without reasoning
and 417-952 with it. All agents fitted on one 48-GB GPU, with peak GPU
memory of 7,368-35,057 MiB for System 1 agents and 12,044-45,532 MiB for
LLM-based agents.

\subsection*{Robustness and reproducibility}
The comparison was independent of the value-matching method. Two values
were considered equivalent when they would have been treated as
equivalent by the server, for example, through its own name-matching
table (Methods). However, the servers did not resolve equivalence in two
cases: stop words, such as ``and'' or ``the'', in database search text,
and model names that an AutoML server passed directly to its underlying
library. To assess the sensitivity of the results to these choices, we
rescored all agents using the alternative matching rule for each case.
Search text was compared exactly as written, while model names were
matched using the library's own name table. The first alternative did
not change any pairwise comparisons. The second changed one of the 324
comparisons from a System 1 advantage to no statistically significant
difference (Extended Data Table 8).

The comparison was also independent of the reasoning mode. For fully
correct requests, System 1 agents outperformed the LLM-based agents in
37 of the 54 pairs with reasoning enabled, while the LLM-based agents
performed better in 6. With reasoning disabled, the corresponding counts
were 36 and 5. When each LLM-based agent was evaluated using its
better-performing mode, selected retrospectively based on the results, a
choice that favored the LLMs, System 1 agent still outperformed the
LLM-based agent in 33 of 54 pairs, compared with 8 in which the
LLM-based agent performed better. The advantage was also not limited to
generated requests. On the 123 prompts taken verbatim from the notebooks
of two published agent systems, System 1 agents achieved mean
tool-calling and argument-filling accuracy at least as high as that of
LLM agents in either reasoning mode (\textbf{Extended Data Table 5}).

Every result can be traced to the released data (Data availability). Replaying the
stored decision-model answers through the released code rebuilds all
12,756 System 1 records of the evaluation set. Rescoring each System 1
record from its recorded call reproduced its stored verdict, and
rescoring the LLM completions reproduced the totals of the run
summaries of all 18 LLM-agent settings. The released scripts can be used to regenerate every figure,
Table 1, Extended Data Tables 3-8 and every statistic in the text.

\section*{Discussion}
Clinical and other sensitive data call for agents that run locally and
do not invent content. Current options force a trade-off. Cloud LLMs are
capable but move data off site. Local LLMs keep data on site but, as we
show, put invented values into 5.5-19.7\% of their tool calls and invent
missing required values for up to 62.1\% of underspecified requests. In
a pipeline, each such value silently changes the analysis and every step
that follows.

System 1 agents remove this trade-off for tool calling. They run on a
local GPU and cannot hallucinate content, because they never generate
text. Every tool, argument and value in a call is chosen from options
that come from the catalogue, the request, the schema or the agent's
policy. We observed no invented content in 10,888 calls. When a required
value is missing, they leave it empty and prompt the user for it with a
template built from the schema. Every decision carries a probability and
every run can be replayed, which makes the behaviour of the agent
auditable.

These properties did not cost accuracy. The two strongest System 1
agents were ahead of every local LLM setting or level with the best one,
mainly because they chose tools and declined more reliably. On average
they were faster and used less energy than LLM-based agents with
reasoning, although Qwen3.5-35B-A3B with reasoning used less energy than
OpenJev-based System 1 agent. At the top of the accuracy range,
OpenJev-based System 1 agent was faster and cheaper than the best LLM
setting. The decision model mattered since the two pplx-decider System 1
models were better tool callers than their base model run as an agent,
whereas decider-35b-a3b was not.

System 1 agents are not error-free. They still make wrong choices among
grounded options. In a limited number of cases, they picked wrong tools,
put written values into the wrong argument and, in 22 of 10,888 calls,
sent grounded text in a form the schema rejects. Checking each option
against the argument's schema, pattern and type, before offering it
would remove the last of these. Requests for which a tool performs the
task but not on the given input were hard for every agent. On argument
filling, five LLM-based agents were not significantly different from the
best System 1 agent.

In conclusion, herein we present a general method to develop System 1
agents. As scientific and clinical tools adopt common interfaces such as
MCP, System 1 agents offer a route to local, reliable and fast
automation of analyses on sensitive data, in which every step is
grounded, scored and auditable.

\section*{Methods}
\subsection*{Decision models}
A decision model maps a state, a question and a list of options to a
probability distribution over the options, in one forward pass and
without sampling. We used six published models from three protocol
families, unchanged and without fine-tuning (\textbf{Extended Data Table
1})\citep{ref14,ref15,ref16,ref17}. Each was served by llama.cpp (build
b11929)\citep{ref23} entirely on one GPU, with one request slot;
every question was its own request.

\emph{Decider family models} (decider-4b, decider-35b-a3b). The prompt
is the family's state-first layout, built as token ids with the
checkpoint's tokenizer: a context block with the state, the question,
the options labelled (A), (B), \ldots{} and an answer prefix. The answer
is read at the final position. The log-probabilities of the option-label
tokens returned by llama-server before sampling, are divided by the
model's fitted temperature and normalized over the valid labels. Up to
ten options use the letters A-J and larger questions use one
single-token label per option, up to 255.

\emph{pplx-decider family models} (pplx-decider-flash,
pplx-decider-27b). The prompt is the checkpoint's chat template around a
fixed system prompt and a user turn with the state, the question and the
options, ending in a marker token. The model's full-attention layers are
non-causal. llama-server runs in embeddings mode and returns the final
hidden state at the marker. The agent applies the model's read-out
matrix (255 × hidden size) and temperature and normalizes over the
options. We verified this path against the reference implementation on
the 4B model.

\emph{Native SystemOne models} (OpenJev, Clef). llama.cpp implements
their prompt, read-out head and temperatures. Its \texttt{/v1/systemone}
endpoint takes the state and one question whose criteria are the option
texts and returns one probability per option. OpenJev takes at most 52
options per question. Larger questions are split into balanced groups.
The model answers within each group and then among the group winners,
and an option's probability is the product of its group winner's final
probability and its probability within the group.

\emph{Determinism and stored answers.} For a given model, hardware and
llama.cpp build, the same state, question, options and kind give the
same distribution. Every answer was stored under the SHA-256 hash of
(model, kind, state, question, options).

\subsection*{Compiler}
The compiler turns one tool's input schema and the request into
questions, sends them to the decision model and assembles the answers
into argument values.

\emph{Argument kinds.} Each property gets a kind from its type, name and
description: enumeration, boolean, number, path, column, free text,
list, array of arrays or object. A union type (anyOf or oneOf, such as
Optional{[}X{]}) is read as its single non-null type.

\emph{Questions.} The state of every question is the request, the tool's
name and description, and any earlier steps. An enumeration is asked as
`Which value should the argument \texttt{x} take?' over its values, plus
`not specified in the request (keep the default)' for an optional
argument. A boolean is asked as `Does the request turn the option
\texttt{x} on or off?' over on, off and, if optional, `not mentioned'.
Numbers, paths, columns and free text use the value question over their
candidates plus `not specified'. List items are picked one at a time
over the remaining candidates plus `no further item'. Written arrays and
dictionaries are offered whole, otherwise an object is filled through a
yes/no question, its keys and one value question per key.

\emph{Candidates.} Candidates come from four sources. (1) The request:
paths, numbers in digits or words, ranges, quoted strings, identifiers
and compact codes, phrases of up to three words for name and search
arguments, written lists and dictionaries, and text after a label. (2)
Forms derived from the request for arguments that take them: unit
conversions, fractions and percentages, class counts, dataset
identifiers, chromosome and allele notations, and paths composed from
written parts. (3) The schema: values a description lists and documented
list items. (4) The environment and earlier steps: the header of a CSV
file another argument points to, and earlier results. Words that name
the tool or its server are not offered.

\emph{Settling and acceptance.} Independent questions are sent together
and settle in a fixed order: free values settle highest probability
first, so a written value goes to the argument that wants it most. A
value fills at most as many arguments as the request writes it. Used
values stay visible but cannot be picked again. When two arguments claim
the same value, the model is asked which argument the value belongs to,
and that answer decides only if its probability reaches 0.7. An optional
argument is set only when the probability on its allowed values reaches
0.7, and a list's first item must reach 0.7. A value equal to the
documented default is not passed. A required argument without a value is
reported as missing, and one answered below 0.5 as uncertain.

\subsection*{Agent}
The agent state consisted of the current user request and all preceding
steps. Tool selection was performed through a multiple-choice question
that included all available tools, each represented by its server and
name followed by the first 220 characters of its description, as well as
a `none of these' option for requests that could not be handled by any
available tool. When the two most probable tools had selection
probabilities within 0.3 of each other, a second question presented the
two tools and their parameter names to determine which tool had
parameters corresponding to the values specified in the request.

A subsequent yes/no question presented the selected tool's full
description and parameter names to assess whether it could fulfil the
request. The agent declined requests when the probability of a positive
answer was below 0.5. In deployment, the agent also declined when its
selection confidence, calculated as the product of the probabilities
assigned to the steps, fell below 0.3. This threshold was not applied
during evaluation, so declines resulted only from selecting `none of
these' or failing the capability check.

When a required argument was missing or uncertain, the agent presented
the user with a fixed prompt containing the argument name and the first
120 characters of its schema description, and used the value provided by
the user. If the request did not specify an output folder, the agent
populated this argument according to a predefined policy. Because the
evaluation was conducted without user interaction, the agent instead
listed missing arguments in its plan. An underspecified request was
considered correctly handled when this list contained exactly the
arguments labelled as missing.

For multi-step requests, each tool call and the first 1,200 characters
of its result were added to the agent state. The agent then used an
additional question to determine whether another step was required.

\subsection*{Development method}
All development was confined to the compiler and agent and was conducted
exclusively on development requests. First, every textual value required
by a tool call was presented to the model as an option, while operations
that could be performed exactly through code were handled by the
compiler. Second, a compiler oracle replaced the decision model with a
component that selected the labelled value whenever it was available.
This provided a measure of which labelled calls could be expressed by
the compiler. Third, each failure was classified as a compiler failure,
a model failure or a labelling error. Compiler failures occurred when a
value was not offered or was offered in an incorrect form. Model
failures occurred when an offered value was not selected. Only compiler
failures were addressed. Fourth, each modification introduced a general
rule governing how requests expressed values or how schemas described
arguments, rather than a tool-specific rule. Its acceptance criteria
were documented before any model was evaluated against it. Fifth, a set
of modifications was retained only if it increased the aggregate number
of fully correct calls across models on the development requests, did
not cause any earlier reference request to fall beyond a predefined
tolerance, and could not be simplified by removing an individual
modification without increasing the aggregate score. Sixth, stored model
responses made repeated evaluations inexpensive and exactly
reproducible. Seventh, the code and configuration were hashed before the
test evaluation.

\subsection*{Tools and catalogues}
A catalogue is a JSON snapshot of each server's tool list: every tool's
name, description and input schema as the server returns it. The tested
157 tools were indexed in two catalogues: (i) the imaging catalogue with
42 tools of 11 MCP servers\citep{ref24}: 2D and 3D image
classification, nnU-Net\citep{ref21}, the Merlin CT foundation
model\citep{ref25}, PyCaret\citep{ref26} classification
and regression, exploratory data analysis, feature importance, model
comparison, PyRadiomics\citep{ref27} and
TotalSegmentator\citep{ref28} (ii) the biomedical catalogue has
115 tools of 7 MCP servers: proteomics-mcp (53 tools),
AlphaGenome\citep{ref29} (22), MSMCP (13), ClinicalTrials.gov
(7), ChEMBL\citep{ref22} (7), Scanpy\citep{ref30} (7) and
TISSUE (6). The AlphaGenome, Scanpy and TISSUE servers come from
Paper2Agent\citep{ref3}, and the ClinicalTrials.gov and ChEMBL
servers from CyanHeads
(\href{https://github.com/cyanheads/chembl-mcp-server}{https://github.com/cyanheads/chembl-mcp-server}).

\subsection*{Prompt Requests, labels and evaluation set}
Prompt requests were designed to resemble explicit, researcher-style
instructions and to cover a range of linguistic and numerical phenomena,
including paraphrasing, distractor values, alternative number formats,
quotation, negation and textual noise. Out-of-scope requests covered
three categories: far, referring to tasks unrelated to the available
tools, near, referring to related tasks for which no available tool
provided the required functionality, and modality, referring to tasks
supported by an available tool but not for the specified input type.

Each request was labelled with the tool whose execution would fulfil it
and no tool for out-of-scope requests. Labels included only argument
values explicitly stated in the request and represented in forms
accepted by the corresponding server. Values not specified in the
request, values requiring external knowledge and values appearing only
in schema examples were excluded. Requests that required a tool but
omitted a mandatory argument were classified as underspecified, with the
missing arguments recorded in their labels.

The evaluation set comprised 2,126 requests: 2,003 custom prompts
(1,179 for the imaging catalogue and 824 for the biomedical catalogue)
and 123 prompts taken verbatim from published research studies (73 from
ReclAIm\citep{ref19} and 50 from mAIstro\citep{ref20}, both for the
imaging catalogue). All 123 published prompts are complete requests.
The evaluation set therefore contains 1,645 complete requests (1,522
custom and 123 published) specifying 6,056 argument values (5,467 and
589), 307 out-of-scope requests that could not be handled by any
available tool and 174 underspecified requests that omitted at least one
required argument; all out-of-scope and underspecified requests are
custom. The 1,952 complete and out-of-scope requests (1,829 custom and
123 published) form the main benchmark, on which tool calling, argument
filling and fully correct requests are measured. The 174 underspecified
requests are evaluated separately.

\subsection*{LLM agents}
Nine open-weight LLMs served as the core reasoning engine of the
LLM-based agents (\textbf{Extended Data Table 2}). The request was the
only user message, with no specific system prompt. Each request received
one completion (seed 20261102, at most 16,384 generated tokens), and no
tool was executed. llama.cpp (build b11929) parsed tool calls with its
native parser. The first call was scored, and an answer without a call
counted as a decline. Each model ran with reasoning on and off through
its chat template. Muse Glimmer 30B has no off setting, so its lowest
level stood for off.

\subsection*{Scoring}
A scoring tool was developed to evaluate every agent. Tool calling was
considered correct when the labelled tool was called for a complete
request or no tool was selected for an out-of-scope request.
Argument-filling accuracy was calculated as the proportion of labelled
values that were correctly provided in calls to the correct tool. A
request was considered fully correct when the correct tool was called,
all labelled values were provided correctly and no required value was
missing, or when an out-of-scope request was declined.

Values were compared as received by the server. Case differences were
ignored, comma-separated lists were compared as sets, numerical values
were compared numerically, and values that a server mapped to the same
entry through its own name-matching table were considered equivalent.
For ClinicalTrials.gov, the scorer applied the server\textquotesingle s
own normalisation procedures, whereas for ChEMBL it used the
server\textquotesingle s assay-type codes. Search text was compared
after removing English stop words\citep{ref31}, irrespective of
word order.

Two matching rules were determined by the scorer. Their sensitivity was
assessed by repeating all pairwise comparisons under alternative rules.
First, search text was compared exactly as provided, without removing
stop words. Second, model names passed by the PyCaret regression server,
which did not have its own name-matching table, were matched using
PyCaret\textquotesingle s table of model identifiers and names rather
than compared as written.

Values not specified in the request were not counted as errors, but were
reported separately. String values corresponding to an
argument\textquotesingle s declared type were converted in the same way
as a schema-validating server would convert them. For underspecified
requests, correct behaviour required calling the correct tool, providing
all stated values correctly and leaving the missing required arguments
unfilled.

\subsection*{Fabricated content}
For every call in the evaluation set, we recorded four types of errors:
(i) the use of a nonexistent tool, defined as a function absent from the
tool catalogue, (ii) the use of a nonexistent argument, defined as an
argument name absent from the schema of the selected tool, (iii) the
inclusion of an invented value and (iv) for underspecified requests, the
filling of a missing required argument.

A value, including a string, number, list item, object key or object
value, was considered grounded if it matched a labelled value, appeared
in the request text (ignoring differences in case, spaces, underscores,
hyphens and slashes), was a number explicitly stated in the request, was
an option offered by the compiler for that argument and request, or
appeared in the tool schema. Booleans, empty values and the output
folder were excluded from this assessment. Values that violated the
schema\textquotesingle s type or pattern constraints were recorded
separately as schema-validity errors.

\subsection*{Local deployment}
All inference ran on one workstation with one GPU, through local
llama.cpp servers. No request, tool schema or answer was sent to an
external service during the runs and the scorer's normalizers also run
locally. Inference was performed on a machine with an NVIDIA RTX PRO
5000 Blackwell GPU (48 GB), 32 CPU threads and 62 GB of memory under
Ubuntu 24.04.

\subsection*{Time and energy}
For each request we recorded wall-clock time, prompt and generated
tokens, GPU energy from the NVML total-energy counter of the whole
board, mean power, utilization and peak memory. Idle power was measured
for 20 s with the model loaded before each run. Energy per fully correct
request is the GPU energy over the 1,952 (1,645 complete and 307
out-of-scope) main requests divided by the number fully correct.

\subsection*{Statistics}
Each System 1 agent was compared with each LLM-based agent in each
reasoning mode (108 pairs per measure). Tool calling and fully correct
requests were compared with the exact two-sided McNemar test on
discordant requests and argument filling with an exact two-sided paired
sign-flip test on per-request differences in right values. P-values were
adjusted with Holm's method over the 108 pairs of each measure, with
significance at an adjusted P \textless{} 0.05. Fully correct requests
are the primary measure; both reasoning modes are co-primary.
Same-backbone comparisons are reported with unadjusted P-values.
Proportions have 95\% Wilson intervals and argument filling has 95\%
percentile intervals from 4,000 bootstrap resamples of requests.

\section*{Data availability}
The requests of the evaluation set with their labels, the two tool
catalogues, the records of every agent, the stored decision-model
answers, the LLM completions and the review of flagged values are
available in the folders \texttt{data} and \texttt{results} of the
repository at \url{https://github.com/eltzanis/SystemOneAgents}.

\section*{Code availability}

The System 1 Agents framework, including the agent implementation, deterministic compiler, evaluation harness, scoring tools, and scripts for reproducing the results presented in this study, is publicly available at \url{https://github.com/eltzanis/SystemOneAgents}.

\section*{Funding}

Funded by the European Union under the Horizon grant agreement No.
101287679. Views and opinions expressed are however those of the
author(s) and do not necessarily reflect those of the European Union nor
European Research Executive Agency (REA). Neither the European Union nor
REA can be held responsible for them.

\section*{Acknowledgements}
None

\section*{Use of large language models}
During the preparation of this manuscript, Claude Opus 5.5 was used for
language polishing and text formatting. The authors reviewed and edited
the content as needed and take full responsibility for the final
publication.

\section*{Author contributions}
\noindent\textbf{Eleftherios Tzanis:} Conceptualization, Software, Resources, Project administration, Methodology, Investigation, Formal analysis, Data curation, Validation, Writing - original draft, Writing - review \& editing.

\noindent\textbf{Michail E. Klontzas:} Supervision, Conceptualization, Project administration, Visualization, Validation, Writing - review \& editing.

\section*{Competing interests}
None

\clearpage
\onecolumn
\begin{figure}[p]
\centering
\includegraphics[width=\linewidth,height=0.68\textheight,keepaspectratio]{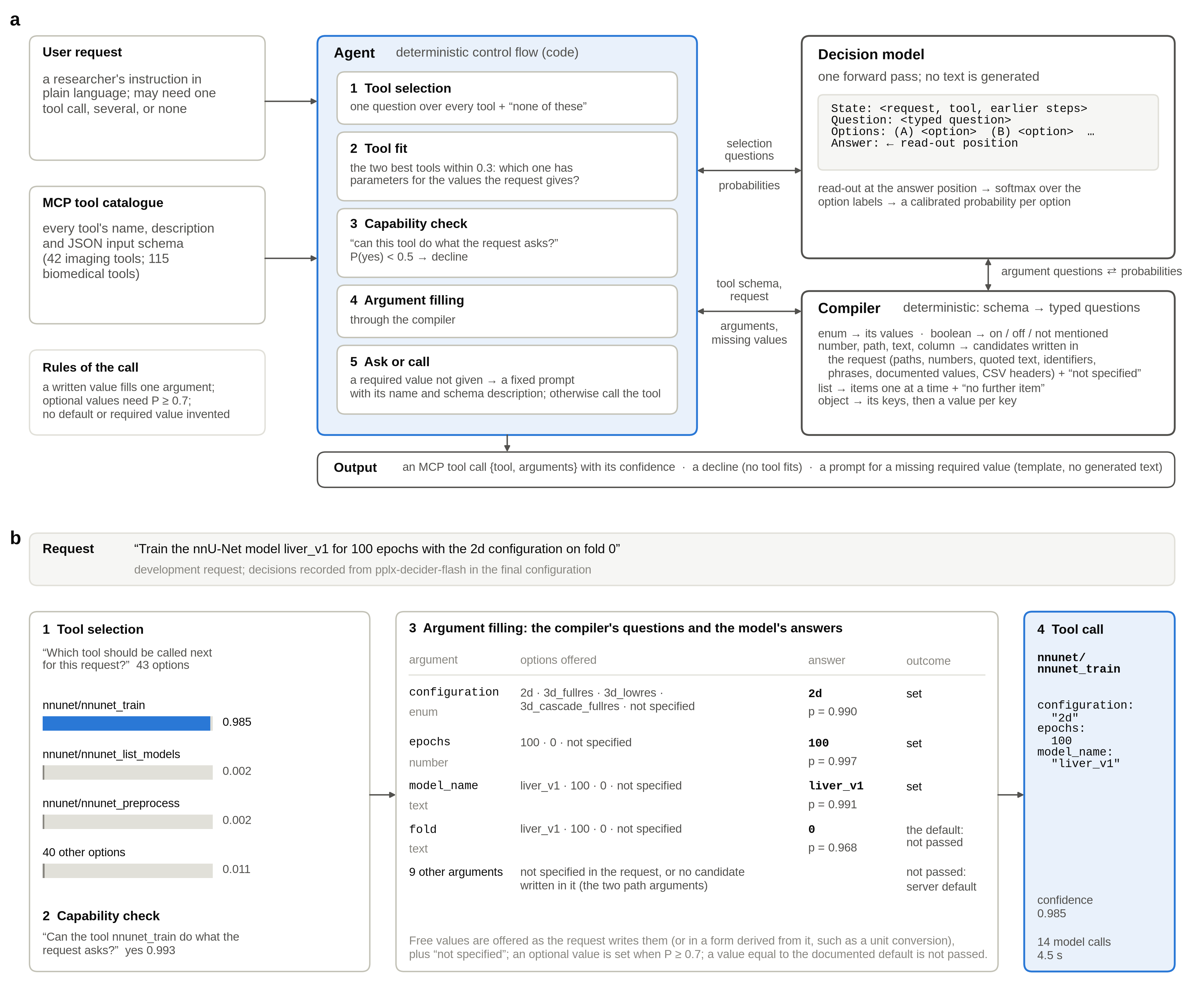}
\caption{\textbf{Architecture of a System 1 agent.} \textbf{a}, Components and data flow. The agent asks the decision model typed questions for tool selection, tool fit and the capability check, and passes the chosen tool and the request to the compiler. The compiler turns the tool's schema into argument questions for the same model and returns the arguments and any missing required values. The agent then prompts the user for a missing required value with a fixed template (the argument's name and schema description) or sends the MCP call. No step generates text. \textbf{b}, One imaging development request with the decisions recorded from pplx-decider-flash: the tool question over 43 options, the capability check, the compiler's questions with the options offered and the model's answers, and the resulting call.}
\label{fig:system1-1}
\end{figure}

\begin{figure}[p]
\centering
\includegraphics[width=\linewidth,height=0.68\textheight,keepaspectratio]{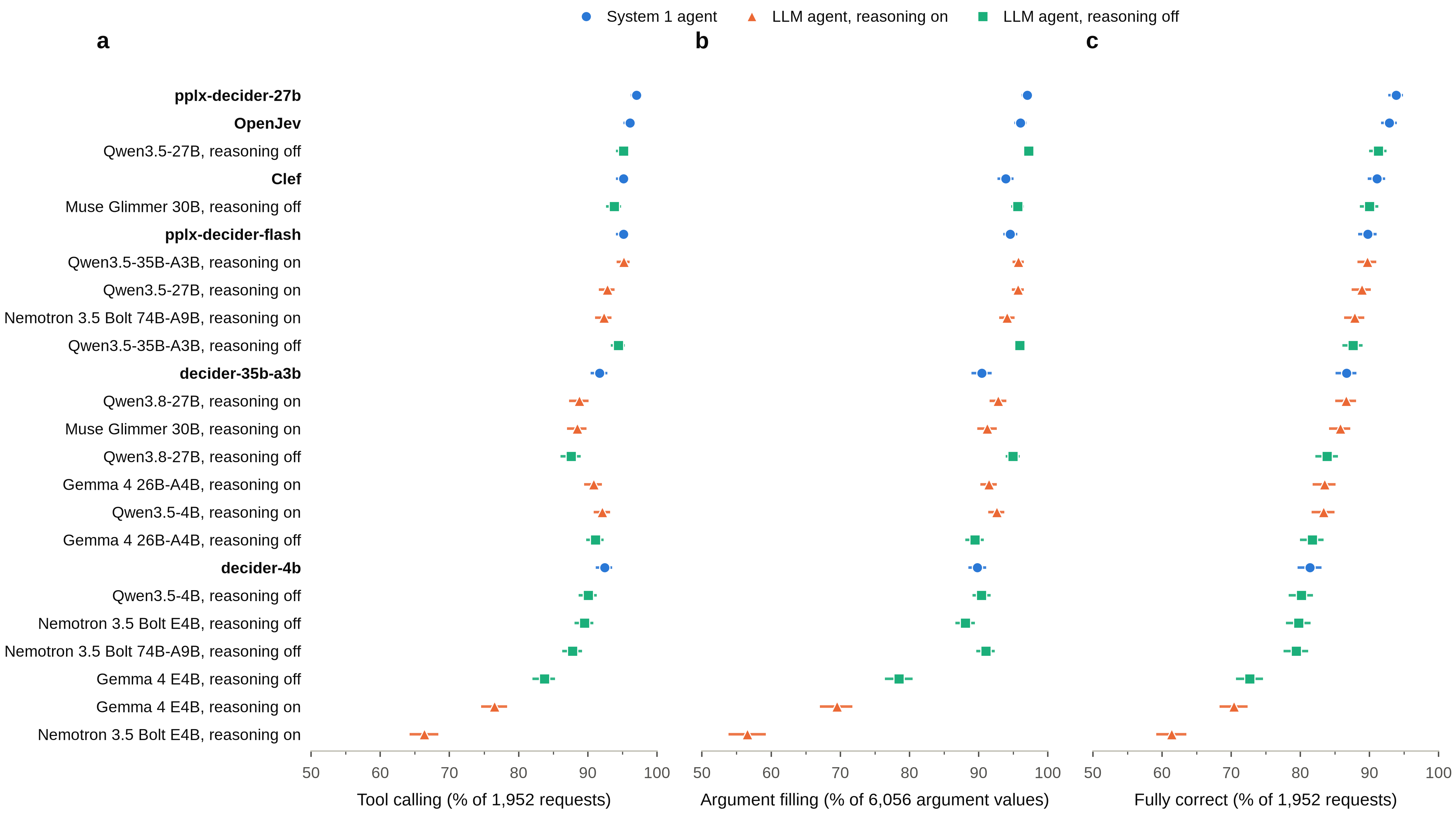}
\caption{\textbf{Accuracy of every agent.} \textbf{a}, Tool calling: the right tool for a complete request or a decline for an out-of-scope request (n = 1,952 requests). \textbf{b}, Argument filling: stated values right in a call to the right tool (n = 6,056 values). \textbf{c}, Fully correct requests: for a complete request, the labelled tool with every stated value right and no required argument left empty; for an out-of-scope request, a decline (n = 1,952). Points show proportions; error bars show 95\% Wilson intervals (\textbf{a},\textbf{c}) and 95\% bootstrap intervals over requests (\textbf{b}). Agents are ordered by fully correct requests; System 1 agents are in bold.}
\label{fig:system1-2}
\end{figure}

\begin{figure}[p]
\centering
\includegraphics[width=\linewidth,height=0.68\textheight,keepaspectratio]{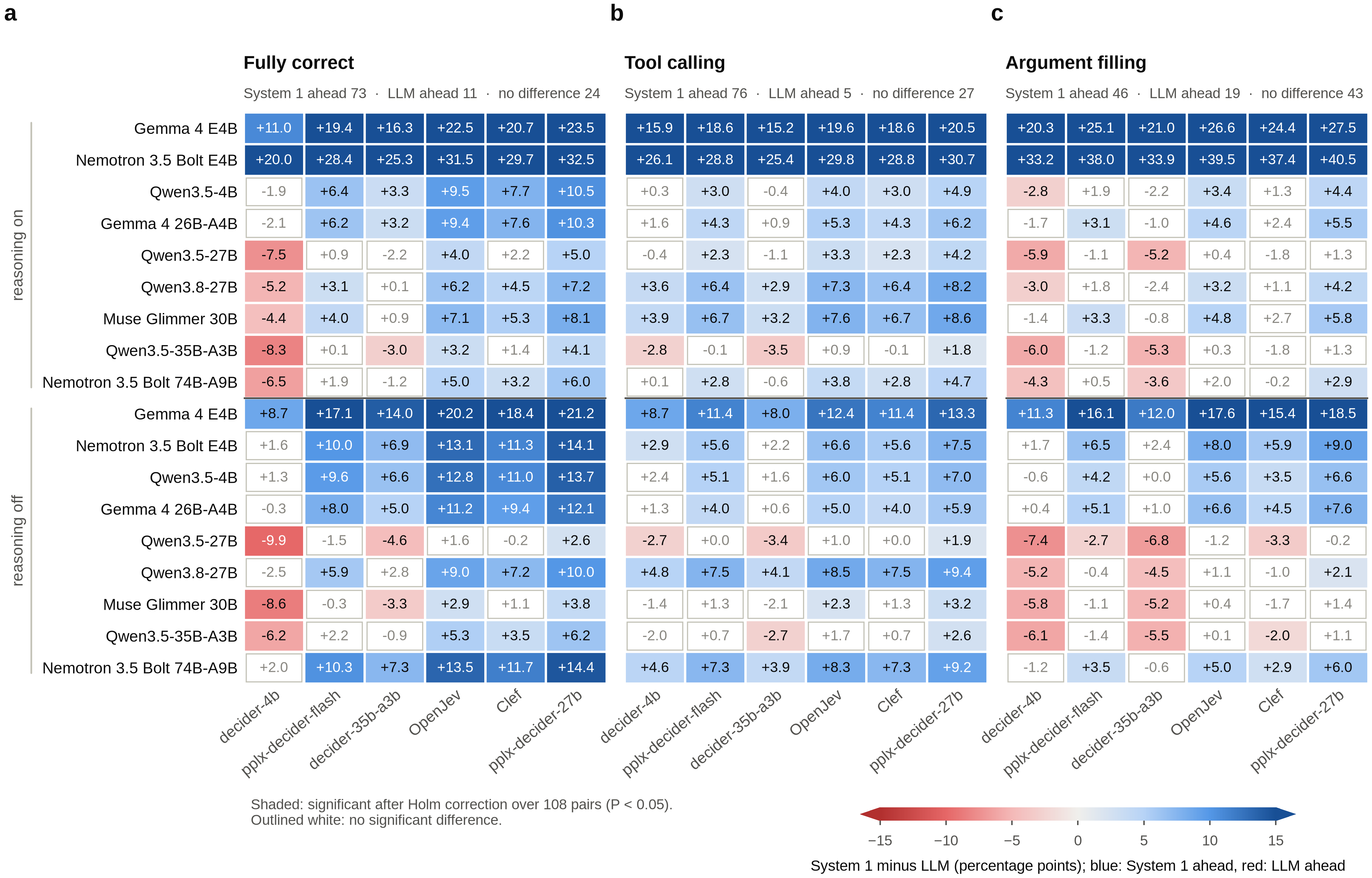}
\caption{\textbf{Pairwise comparison of System 1 and LLM agents.} Difference in percentage points (System 1 agent minus LLM-based agent) for each System 1 agent (columns) and each LLM-based agent in each reasoning mode (rows), for \textbf{a}, fully correct requests, \textbf{b}, tool calling and \textbf{c}, argument filling. Shaded cells are significant after Holm correction over the 108 pairs of each measure (adjusted P \textless{} 0.05; exact McNemar test in \textbf{a},\textbf{b}, exact paired sign-flip test in \textbf{c}); outlined cells are not significant. Colours are clipped at ±15 points. Above each panel: pairs with System 1 ahead, LLM ahead and no significant difference.}
\label{fig:system1-3}
\end{figure}

\begin{figure}[p]
\centering
\includegraphics[width=\linewidth,height=0.68\textheight,keepaspectratio]{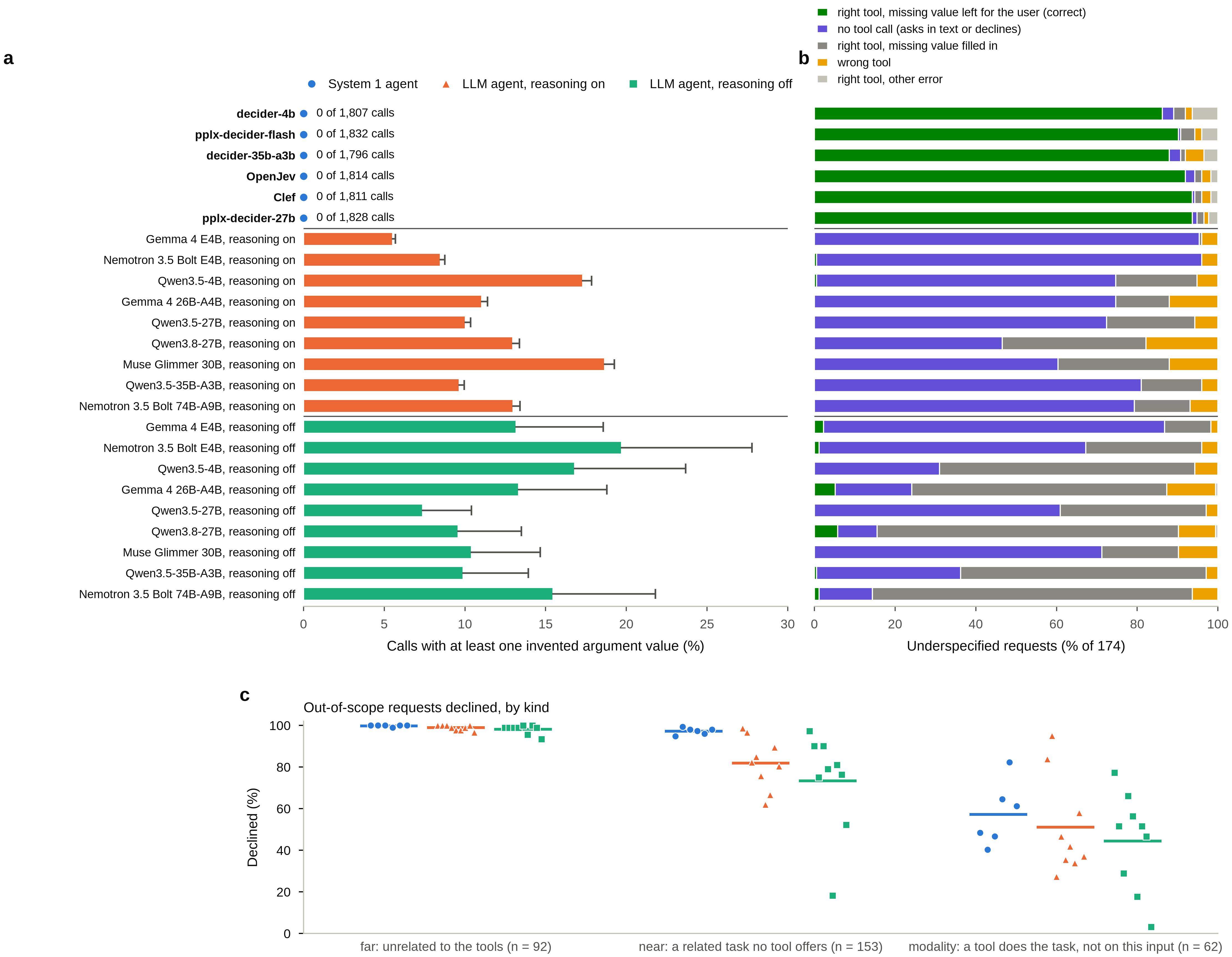}
\caption{\textbf{Reliability evaluation.} \textbf{a}, Calls with at least one invented argument value, per agent. Bars show the lower bound and whiskers the upper bound after correcting the automatic check by the precision measured in an independent review (n = 200 flagged values). System 1 agents had none in 10,888 calls. \textbf{b}, Outcomes on the 174 underspecified requests. The correct behaviour is the right tool with the missing required value left for the user; for System 1 agents, every filled value was written elsewhere in the request. \textbf{c}, Out-of-scope requests declined, by kind: far (n = 92), near (n = 153) and modality (n = 62). Points are agents; bars mark the mean of each kind of agent.}
\label{fig:system1-4}
\end{figure}

\begin{figure}[p]
\centering
\includegraphics[width=\linewidth,height=0.68\textheight,keepaspectratio]{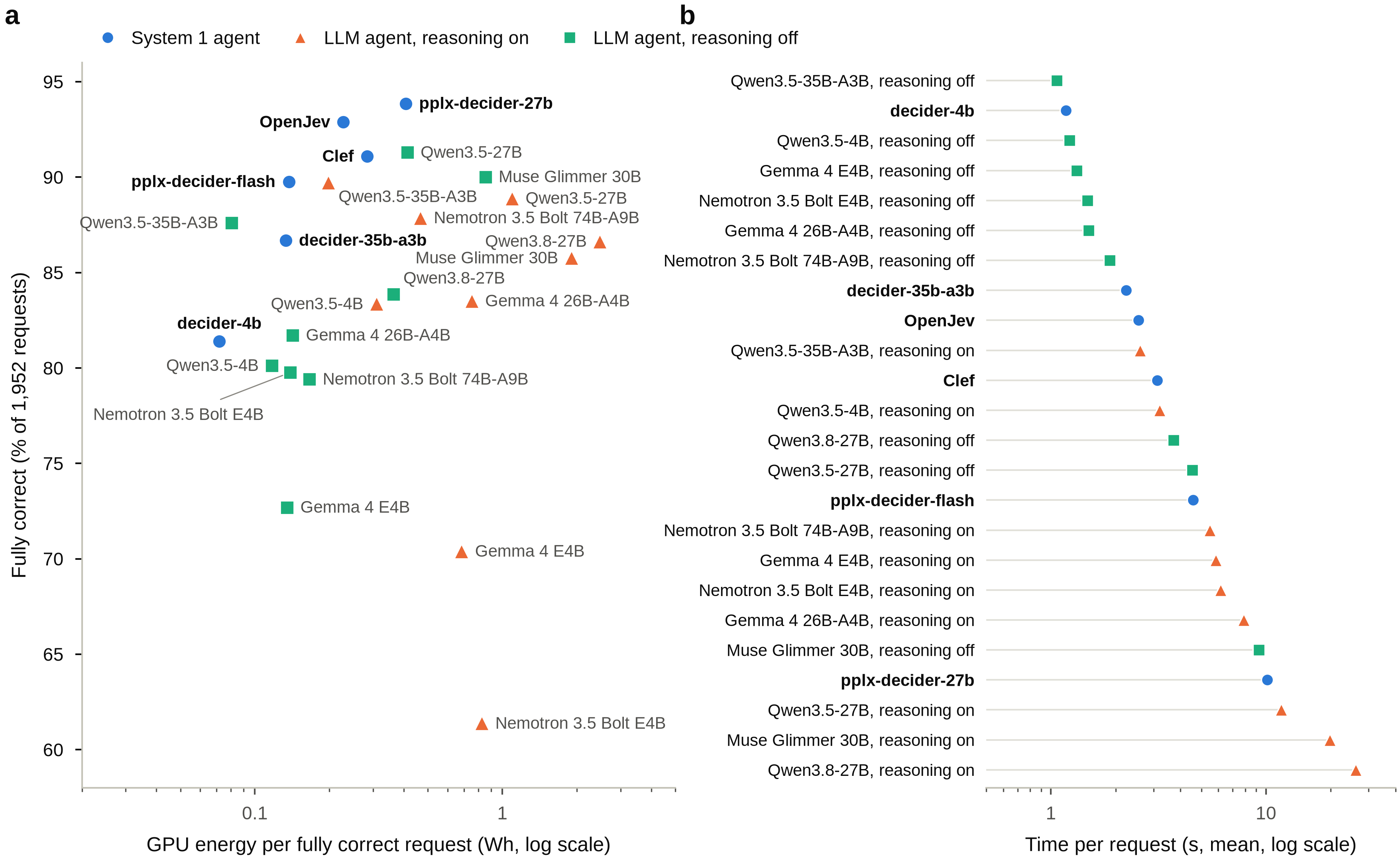}
\caption{\textbf{Efficiency evaluation.} \textbf{a}, Fully correct requests against GPU energy per fully correct request (log scale). Marker shape and colour give the kind of agent and, for LLM agents, the reasoning mode; every agent is labelled. Extended Data Table 6 gives the values. \textbf{b}, Mean time per request, all agents ranked (log scale).}
\label{fig:system1-5}
\end{figure}

\clearpage
\section*{Tables}
\begingroup
\footnotesize
\setlength{\tabcolsep}{2.1pt}
\setlength{\LTpre}{9pt}
\setlength{\LTpost}{14pt}

\noindent\textbf{Table 1. System 1 agents and LLM agents compared by category.}
Tool calling: the labelled tool called for a complete request, or a
decline for an out-of-scope request. Argument filling: stated values
right in a call to the right tool. Fully correct: for a complete
request, the labelled tool called with every stated value right and no
required argument left empty; for an out-of-scope request, a decline.
Ranges over the six System 1 agents and the 18 LLM-agent settings (nine
models, reasoning on and off), the 108-pair tallies for the accuracy
measures (System 1 significantly ahead / LLM significantly ahead / no
significant difference; Holm-adjusted P \textless{} 0.05), and three
leading agents: the most accurate System 1 agent (pplx-decider-27b), a
fast System 1 agent (OpenJev) and the most accurate LLM setting
(Qwen3.5-27B, reasoning off). \textsuperscript{a}LLM values are the
precision-adjusted lower bound. \textsuperscript{b}For System 1 agents,
tokens the server computes for the read-out; no text is generated.
\textsuperscript{c}Invented: a value written nowhere in the request.

\begin{longtable}[]{@{}
  >{\raggedright\arraybackslash}p{(\columnwidth - 14\tabcolsep) * \real{0.1243}}
  >{\raggedright\arraybackslash}p{(\columnwidth - 14\tabcolsep) * \real{0.1280}}
  >{\raggedright\arraybackslash}p{(\columnwidth - 14\tabcolsep) * \real{0.1227}}
  >{\raggedright\arraybackslash}p{(\columnwidth - 14\tabcolsep) * \real{0.1250}}
  >{\raggedright\arraybackslash}p{(\columnwidth - 14\tabcolsep) * \real{0.1365}}
  >{\raggedright\arraybackslash}p{(\columnwidth - 14\tabcolsep) * \real{0.1135}}
  >{\raggedright\arraybackslash}p{(\columnwidth - 14\tabcolsep) * \real{0.1250}}
  >{\raggedright\arraybackslash}p{(\columnwidth - 14\tabcolsep) * \real{0.1250}}@{}}
\toprule\noalign{}
\begin{minipage}[b]{\linewidth}\raggedright
category
\end{minipage} & \begin{minipage}[b]{\linewidth}\raggedright
measure
\end{minipage} & \begin{minipage}[b]{\linewidth}\raggedright
System 1 (six agents)
\end{minipage} & \begin{minipage}[b]{\linewidth}\raggedright
LLM (18 agents)
\end{minipage} & \begin{minipage}[b]{\linewidth}\raggedright
pairs
\end{minipage} & \begin{minipage}[b]{\linewidth}\raggedright
pplx-decider-27b
\end{minipage} & \begin{minipage}[b]{\linewidth}\raggedright
OpenJev
\end{minipage} & \begin{minipage}[b]{\linewidth}\raggedright
Qwen3.5-27B, reasoning off
\end{minipage} \\
\midrule\noalign{}
\endhead
\bottomrule\noalign{}
\endlastfoot
Accuracy & Fully correct requests (\% of 1,952) & 81.4--93.9 &
61.4--91.3 & 73 / 11 / 24 & 93.9 & 92.9 & 91.3 \\
Accuracy & Tool calling (\% of 1,952) & 91.7--97.0 & 66.3--95.2 & 76 / 5
/ 27 & 97.0 & 96.1 & 95.1 \\
Accuracy & Argument filling (\% of 6,056 values) & 89.8--97.0 &
56.6--97.2 & 46 / 19 / 43 & 97.0 & 96.1 & 97.2 \\
Reliability & Calls with invented content (\% of
calls)\textsuperscript{a} & 0.0 & 5.5--19.7 & --- & 0.0 & 0.0 & 7.4 \\
Reliability & Calls with argument names absent from the schema (n) & 0 &
0--53 & --- & 0 & 0 & 0 \\
Reliability & Underspecified requests handled correctly (\% of 174) &
86.2--93.7 & 0.0--5.7 & --- & 93.7 & 92.0 & 0.0 \\
Reliability & Underspecified requests with an invented required value
(\% of 174)\textsuperscript{c} & 0.0 & 0.0--62.1 & --- & 0.0 & 0.0 & 27.6 \\
Reliability & Out-of-scope requests declined (\% of 307) & 87.0--94.5 &
41.4--97.4 & --- & 91.2 & 91.2 & 80.8 \\
Reliability & Complete requests left without a call (\% of 1,645) &
1.0--3.3 & 0.0--36.5 & --- & 1.0 & 1.7 & 0.7 \\
Reliability & Calls with a schema-invalid value (\% of calls) & 0.0--0.4
& 0.0--0.8 & --- & 0.0 & 0.1 & 0.1 \\
Efficiency & GPU energy per fully correct request (Wh) & 0.072--0.407 &
0.080--2.481 & --- & 0.407 & 0.228 & 0.413 \\
Efficiency & Time per request (s) & 1.2--10.1 & 1.1--26.1 & --- & 10.1 &
2.6 & 4.5 \\
Efficiency & Generated tokens per request\textsuperscript{b} & 0--15 &
123--952 & --- & 0 & 0 & 156 \\
Efficiency & Peak GPU memory (MiB) & 7,368--35,057 & 12,044--45,532 &
--- & 35,057 & 21,852 & 36,002 \\
\end{longtable}

\clearpage
\section*{Extended Data}
\noindent\textbf{Extended Data Table 1. Decision models.} Context: the server's
context window in tokens.

\begin{longtable}[]{@{}
  >{\raggedright\arraybackslash}p{(\columnwidth - 10\tabcolsep) * \real{0.1667}}
  >{\raggedright\arraybackslash}p{(\columnwidth - 10\tabcolsep) * \real{0.1667}}
  >{\raggedright\arraybackslash}p{(\columnwidth - 10\tabcolsep) * \real{0.1667}}
  >{\raggedright\arraybackslash}p{(\columnwidth - 10\tabcolsep) * \real{0.1667}}
  >{\raggedright\arraybackslash}p{(\columnwidth - 10\tabcolsep) * \real{0.1667}}
  >{\raggedright\arraybackslash}p{(\columnwidth - 10\tabcolsep) * \real{0.1667}}@{}}
\toprule\noalign{}
\begin{minipage}[b]{\linewidth}\raggedright
model
\end{minipage} & \begin{minipage}[b]{\linewidth}\raggedright
source
\end{minipage} & \begin{minipage}[b]{\linewidth}\raggedright
base model
\end{minipage} & \begin{minipage}[b]{\linewidth}\raggedright
file
\end{minipage} & \begin{minipage}[b]{\linewidth}\raggedright
protocol
\end{minipage} & \begin{minipage}[b]{\linewidth}\raggedright
context
\end{minipage} \\
\midrule\noalign{}
\endhead
\bottomrule\noalign{}
\endlastfoot
decider-4b & Mapika, Mapika/decider-4b-GGUF & Qwen3.5-4B & Q8\_0, 4.5 GB
& decider & 16,384 \\
pplx-decider-flash v1.2 & Perplexity,
perplexity-ai/pplx-decider-flash-gguf-v1.2 & Qwen3.5-4B & Q8\_0, 5.2 GB
& pplx-decider & 5,632 \\
decider-35b-a3b & Mapika, Mapika/decider-35b-a3b (GGUF by mradermacher)
& Qwen3.5-35B-A3B (3B active) & Q4\_K\_M, 21.2 GB & decider & 16,384 \\
OpenJev & openjev/openjev (GGUF by ggml-org) & 27B, Qwen3.5 architecture
& Q4\_K\_M, 19.4 GB & native SystemOne & 8,192 \\
Clef & Cloudflare (GGUF by ggml-org) & Qwen3.5-27B with a joint schema
head & Q8\_0, 28.7 GB & native SystemOne & 8,192 \\
pplx-decider-27b v1.2 & Perplexity, perplexity-ai/pplx-decider-v1.2 &
Qwen3.5-27B & Q8\_0, 28.6 GB, converted with llama.cpp & pplx-decider &
5,632 \\
\end{longtable}

\noindent\textbf{Extended Data Table 2. LLM-based agents.}

\begin{longtable}[]{@{}
  >{\raggedright\arraybackslash}p{(\columnwidth - 6\tabcolsep) * \real{0.2500}}
  >{\raggedright\arraybackslash}p{(\columnwidth - 6\tabcolsep) * \real{0.2500}}
  >{\raggedright\arraybackslash}p{(\columnwidth - 6\tabcolsep) * \real{0.2500}}
  >{\raggedright\arraybackslash}p{(\columnwidth - 6\tabcolsep) * \real{0.2500}}@{}}
\toprule\noalign{}
\begin{minipage}[b]{\linewidth}\raggedright
model
\end{minipage} & \begin{minipage}[b]{\linewidth}\raggedright
parameters
\end{minipage} & \begin{minipage}[b]{\linewidth}\raggedright
file
\end{minipage} & \begin{minipage}[b]{\linewidth}\raggedright
role
\end{minipage} \\
\midrule\noalign{}
\endhead
\bottomrule\noalign{}
\endlastfoot
NVIDIA Nemotron 3.5 Bolt E4B & 8B (4B effective) & Unsloth UD-Q8\_K\_XL
& chosen \\
NVIDIA Nemotron 3.5 Bolt 74B-A9B & 74B, 9B active & NVIDIA 24 GB release
& chosen \\
Qwen3.8-27B & 27B & ggml-org Q8\_0 & chosen \\
Meta Muse Glimmer 30B & 30B & Unsloth Q8\_0 & chosen \\
Google Gemma 4 E4B (instruction-tuned) & 8B (4B effective) & ggml-org
Q8\_0 & chosen \\
Google Gemma 4 26B-A4B (instruction-tuned) & 26B, 4B active & ggml-org
Q8\_0 & chosen \\
Qwen3.5-4B & 4B & Unsloth Q8\_0 & base of decider-4b and
pplx-decider-flash \\
Qwen3.5-27B & 27B & Unsloth Q8\_0 & base of Clef and pplx-decider-27b \\
Qwen3.5-35B-A3B & 35B, 3B active & Unsloth Q4\_K\_M & base of
decider-35b-a3b \\
\end{longtable}

\noindent\textbf{Extended Data Table 3. Accuracy of every agent.} Counts with
percentages, and 95\% Wilson intervals for tool calling and fully
correct requests. Tool calling: the labelled tool
called for a complete request, or a decline for an out-of-scope request.
Argument filling: stated values right in a call to the right tool. Fully
correct: for a complete request, the labelled tool called with every
stated value right and no required argument left empty; for an
out-of-scope request, a decline.

\begin{longtable}[]{@{}
  >{\raggedright\arraybackslash}p{(\columnwidth - 12\tabcolsep) * \real{0.1429}}
  >{\raggedright\arraybackslash}p{(\columnwidth - 12\tabcolsep) * \real{0.1429}}
  >{\raggedright\arraybackslash}p{(\columnwidth - 12\tabcolsep) * \real{0.1428}}
  >{\raggedright\arraybackslash}p{(\columnwidth - 12\tabcolsep) * \real{0.1428}}
  >{\raggedright\arraybackslash}p{(\columnwidth - 12\tabcolsep) * \real{0.1428}}
  >{\raggedright\arraybackslash}p{(\columnwidth - 12\tabcolsep) * \real{0.1428}}
  >{\raggedright\arraybackslash}p{(\columnwidth - 12\tabcolsep) * \real{0.1428}}@{}}
\toprule\noalign{}
\begin{minipage}[b]{\linewidth}\raggedright
agent
\end{minipage} & \begin{minipage}[b]{\linewidth}\raggedright
kind
\end{minipage} & \begin{minipage}[b]{\linewidth}\raggedright
tool calling, of 1,952
\end{minipage} & \begin{minipage}[b]{\linewidth}\raggedright
right tool, of 1,645
\end{minipage} & \begin{minipage}[b]{\linewidth}\raggedright
declined, of 307
\end{minipage} & \begin{minipage}[b]{\linewidth}\raggedright
argument values, of 6,056
\end{minipage} & \begin{minipage}[b]{\linewidth}\raggedright
fully correct, of 1,952
\end{minipage} \\
\midrule\noalign{}
\endhead
\bottomrule\noalign{}
\endlastfoot
decider-4b & System 1 & 1,804 (92.4; 91.2--93.5) & 1,537 & 267 & 5,438
(89.8) & 1,589 (81.4; 79.6--83.1) \\
pplx-decider-flash & System 1 & 1,857 (95.1; 94.1--96.0) & 1,588 & 269 &
5,727 (94.6) & 1,752 (89.8; 88.3--91.0) \\
decider-35b-a3b & System 1 & 1,790 (91.7; 90.4--92.8) & 1,519 & 271 &
5,478 (90.5) & 1,692 (86.7; 85.1--88.1) \\
OpenJev & System 1 & 1,876 (96.1; 95.2--96.9) & 1,596 & 280 & 5,817
(96.1) & 1,813 (92.9; 91.7--93.9) \\
Clef & System 1 & 1,857 (95.1; 94.1--96.0) & 1,567 & 290 & 5,688 (93.9)
& 1,778 (91.1; 89.7--92.3) \\
pplx-decider-27b & System 1 & 1,894 (97.0; 96.2--97.7) & 1,614 & 280 &
5,876 (97.0) & 1,832 (93.9; 92.7--94.8) \\
lexical baseline & no model & 962 (49.3; 47.1--51.5) & 944 & 18 & 1,244
(20.5) & 212 (10.9; 9.6--12.3) \\
Gemma 4 E4B & LLM, reasoning on & 1,493 (76.5; 74.6--78.3) & 1,198 & 295
& 4,209 (69.5) & 1,374 (70.4; 68.3--72.4) \\
Nemotron 3.5 Bolt E4B & LLM, reasoning on & 1,295 (66.3; 64.2--68.4) &
996 & 299 & 3,426 (56.6) & 1,198 (61.4; 59.2--63.5) \\
Qwen3.5-4B & LLM, reasoning on & 1,798 (92.1; 90.8--93.2) & 1,563 & 235
& 5,609 (92.6) & 1,627 (83.4; 81.6--84.9) \\
Gemma 4 26B-A4B & LLM, reasoning on & 1,773 (90.8; 89.5--92.0) & 1,523 &
250 & 5,540 (91.5) & 1,630 (83.5; 81.8--85.1) \\
Qwen3.5-27B & LLM, reasoning on & 1,812 (92.8; 91.6--93.9) & 1,584 & 228
& 5,795 (95.7) & 1,735 (88.9; 87.4--90.2) \\
Qwen3.8-27B & LLM, reasoning on & 1,733 (88.8; 87.3--90.1) & 1,522 & 211
& 5,621 (92.8) & 1,691 (86.6; 85.0--88.1) \\
Muse Glimmer 30B & LLM, reasoning on & 1,727 (88.5; 87.0--89.8) & 1,513
& 214 & 5,525 (91.2) & 1,674 (85.8; 84.1--87.2) \\
Qwen3.5-35B-A3B & LLM, reasoning on & 1,858 (95.2; 94.1--96.0) & 1,593 &
265 & 5,799 (95.8) & 1,751 (89.7; 88.3--91.0) \\
Nemotron 3.5 Bolt 74B-A9B & LLM, reasoning on & 1,802 (92.3; 91.0--93.4)
& 1,567 & 235 & 5,698 (94.1) & 1,715 (87.9; 86.3--89.2) \\
Gemma 4 E4B & LLM, reasoning off & 1,634 (83.7; 82.0--85.3) & 1,346 &
288 & 4,753 (78.5) & 1,419 (72.7; 70.7--74.6) \\
Nemotron 3.5 Bolt E4B & LLM, reasoning off & 1,747 (89.5; 88.1--90.8) &
1,486 & 261 & 5,333 (88.1) & 1,557 (79.8; 77.9--81.5) \\
Qwen3.5-4B & LLM, reasoning off & 1,758 (90.1; 88.7--91.3) & 1,534 & 224
& 5,475 (90.4) & 1,564 (80.1; 78.3--81.8) \\
Gemma 4 26B-A4B & LLM, reasoning off & 1,778 (91.1; 89.7--92.3) & 1,508
& 270 & 5,416 (89.4) & 1,595 (81.7; 79.9--83.4) \\
Qwen3.5-27B & LLM, reasoning off & 1,857 (95.1; 94.1--96.0) & 1,609 &
248 & 5,888 (97.2) & 1,782 (91.3; 90.0--92.5) \\
Qwen3.8-27B & LLM, reasoning off & 1,710 (87.6; 86.1--89.0) & 1,583 &
127 & 5,751 (95.0) & 1,637 (83.9; 82.2--85.4) \\
Muse Glimmer 30B & LLM, reasoning off & 1,831 (93.8; 92.6--94.8) & 1,583
& 248 & 5,792 (95.6) & 1,757 (90.0; 88.6--91.3) \\
Qwen3.5-35B-A3B & LLM, reasoning off & 1,843 (94.4; 93.3--95.4) & 1,606
& 237 & 5,809 (95.9) & 1,710 (87.6; 86.1--89.0) \\
Nemotron 3.5 Bolt 74B-A9B & LLM, reasoning off & 1,714 (87.8;
86.3--89.2) & 1,546 & 168 & 5,513 (91.0) & 1,550 (79.4; 77.6--81.1) \\
\end{longtable}

\noindent\textbf{Extended Data Table 4. Same backbone agent comparison.} Each
of the five System 1 agents whose decision model is built on one of the
evaluated LLMs, against that LLM: difference in percentage
points, System 1 agents minus LLM-based agents (unadjusted P). Fully
correct: for a complete request, the labelled tool called with every
stated value right and no required argument left empty; for an
out-of-scope request, a decline.

\begin{longtable}[]{@{}
  >{\raggedright\arraybackslash}p{(\columnwidth - 8\tabcolsep) * \real{0.2000}}
  >{\raggedright\arraybackslash}p{(\columnwidth - 8\tabcolsep) * \real{0.2000}}
  >{\raggedright\arraybackslash}p{(\columnwidth - 8\tabcolsep) * \real{0.2000}}
  >{\raggedright\arraybackslash}p{(\columnwidth - 8\tabcolsep) * \real{0.2000}}
  >{\raggedright\arraybackslash}p{(\columnwidth - 8\tabcolsep) * \real{0.2000}}@{}}
\toprule\noalign{}
\begin{minipage}[b]{\linewidth}\raggedright
System 1 model
\end{minipage} & \begin{minipage}[b]{\linewidth}\raggedright
base model as LLM agent
\end{minipage} & \begin{minipage}[b]{\linewidth}\raggedright
tool calling
\end{minipage} & \begin{minipage}[b]{\linewidth}\raggedright
argument filling
\end{minipage} & \begin{minipage}[b]{\linewidth}\raggedright
fully correct
\end{minipage} \\
\midrule\noalign{}
\endhead
\bottomrule\noalign{}
\endlastfoot
decider-4b & Qwen3.5-4B, reasoning on & +0.3 (0.72) & −2.8
(\textless0.001) & −1.9 (0.08) \\
decider-4b & Qwen3.5-4B, reasoning off & +2.4 (0.003) & −0.6 (0.47) &
+1.3 (0.27) \\
pplx-decider-flash & Qwen3.5-4B, reasoning on & +3.0 (\textless0.001) &
+1.9 (0.0064) & +6.4 (\textless0.001) \\
pplx-decider-flash & Qwen3.5-4B, reasoning off & +5.1 (\textless0.001) &
+4.2 (\textless0.001) & +9.6 (\textless0.001) \\
Clef & Qwen3.5-27B, reasoning on & +2.3 (0.0017) & −1.8 (0.011) & +2.2
(0.019) \\
Clef & Qwen3.5-27B, reasoning off & 0.0 (1.0) & −3.3 (\textless0.001) &
−0.2 (0.86) \\
pplx-decider-27b & Qwen3.5-27B, reasoning on & +4.2 (\textless0.001) &
+1.3 (0.009) & +5.0 (\textless0.001) \\
pplx-decider-27b & Qwen3.5-27B, reasoning off & +1.9 (\textless0.001) &
−0.2 (0.71) & +2.6 (\textless0.001) \\
decider-35b-a3b & Qwen3.5-35B-A3B, reasoning on & −3.5 (\textless0.001)
& −5.3 (\textless0.001) & −3.0 (0.0013) \\
decider-35b-a3b & Qwen3.5-35B-A3B, reasoning off & −2.7 (\textless0.001)
& −5.5 (\textless0.001) & −0.9 (0.36) \\
\end{longtable}

\noindent\textbf{Extended Data Table 5. Accuracy by subcategory.} Mean over the
agents of each kind (\%), System 1 / LLM reasoning on / LLM reasoning
off.

\begin{longtable}[]{@{}
  >{\raggedright\arraybackslash}p{(\columnwidth - 8\tabcolsep) * \real{0.2000}}
  >{\raggedright\arraybackslash}p{(\columnwidth - 8\tabcolsep) * \real{0.2000}}
  >{\raggedright\arraybackslash}p{(\columnwidth - 8\tabcolsep) * \real{0.2000}}
  >{\raggedright\arraybackslash}p{(\columnwidth - 8\tabcolsep) * \real{0.2000}}
  >{\raggedright\arraybackslash}p{(\columnwidth - 8\tabcolsep) * \real{0.2000}}@{}}
\toprule\noalign{}
\begin{minipage}[b]{\linewidth}\raggedright
subcategory
\end{minipage} & \begin{minipage}[b]{\linewidth}\raggedright
requests
\end{minipage} & \begin{minipage}[b]{\linewidth}\raggedright
values
\end{minipage} & \begin{minipage}[b]{\linewidth}\raggedright
tool calling
\end{minipage} & \begin{minipage}[b]{\linewidth}\raggedright
argument filling
\end{minipage} \\
\midrule\noalign{}
\endhead
\bottomrule\noalign{}
\endlastfoot
source: generated requests & 1,829 & 5,467 & 94.3 / 86.5 / 89.8 & 93.1 /
86.1 / 90.8 \\
source: mAIstro notebooks & 50 & 214 & 100.0 / 96.4 / 97.1 & 98.2 / 92.5
/ 92.6 \\
source: ReclAIm notebooks & 73 & 375 & 98.9 / 93.3 / 98.8 & 98.8 / 91.7
/ 96.9 \\
tools: AutoML (PyCaret) & 112 & 374 & 99.9 / 86.8 / 98.0 & 98.5 / 84.6 /
94.3 \\
tools: chemistry (ChEMBL) & 32 & 61 & 93.8 / 95.1 / 97.2 & 89.6 / 93.6 /
91.6 \\
tools: clinical trials & 41 & 114 & 89.8 / 98.6 / 96.5 & 82.9 / 95.4 /
88.2 \\
tools: 2D image classification & 322 & 1,679 & 98.7 / 89.5 / 96.0 & 97.8
/ 88.8 / 94.6 \\
tools: 3D image classification & 172 & 771 & 98.8 / 81.5 / 93.5 & 98.2 /
80.4 / 91.2 \\
tools: exploratory data analysis & 35 & 110 & 95.7 / 91.4 / 100.0 & 95.9
/ 91.5 / 99.6 \\
tools: model evaluation & 41 & 132 & 99.6 / 99.7 / 99.7 & 98.5 / 99.3 /
98.8 \\
tools: feature selection & 30 & 125 & 98.9 / 89.3 / 98.5 & 96.1 / 87.0 /
94.6 \\
tools: genomics (AlphaGenome) & 135 & 618 & 91.2 / 91.7 / 95.0 & 84.5 /
89.6 / 91.5 \\
tools: Merlin & 66 & 185 & 97.5 / 96.6 / 100.0 & 97.4 / 95.9 / 99.2 \\
tools: metabolomics (MSMCP) & 46 & 92 & 91.7 / 97.1 / 96.4 & 88.6 / 97.8
/ 96.6 \\
tools: nnU-Net & 116 & 403 & 94.7 / 78.6 / 82.5 & 93.8 / 80.5 / 83.0 \\
tools: proteomics & 281 & 792 & 89.4 / 82.8 / 84.9 & 83.6 / 78.1 /
79.2 \\
tools: radiomics & 38 & 165 & 99.1 / 99.1 / 98.8 & 96.0 / 97.7 / 96.6 \\
tools: segmentation & 41 & 131 & 99.2 / 95.9 / 98.1 & 98.6 / 94.2 /
94.8 \\
tools: single cell (Scanpy) & 42 & 106 & 96.8 / 91.3 / 97.6 & 95.9 /
91.5 / 97.3 \\
tools: spatial (TISSUE) & 35 & 136 & 91.4 / 81.6 / 90.8 & 92.4 / 79.7 /
90.3 \\
tools: workflow & 60 & 62 & 96.4 / 91.5 / 90.4 & 95.4 / 96.2 / 94.6 \\
out of scope: near & 153 & --- & 97.3 / 81.9 / 73.3 & --- \\
out of scope: far & 92 & --- & 99.8 / 98.9 / 98.2 & --- \\
out of scope: modality & 62 & --- & 57.3 / 51.1 / 44.4 & --- \\
phenomenon: several arguments & 1,123 & 4,687 & 96.0 / 88.3 / 94.1 &
94.3 / 86.7 / 91.5 \\
phenomenon: distractor values & 938 & 3,970 & 96.4 / 89.8 / 94.4 & 94.5
/ 88.2 / 92.4 \\
phenomenon: numbers & 778 & 3,538 & 95.3 / 87.8 / 93.2 & 93.7 / 86.2 /
90.9 \\
phenomenon: quoting & 356 & 1,430 & 97.3 / 90.9 / 95.6 & 96.6 / 89.4 /
93.8 \\
phenomenon: minimal wording & 355 & 687 & 96.5 / 87.9 / 90.9 & 96.4 /
88.1 / 91.5 \\
phenomenon: path forms & 314 & 1,013 & 94.7 / 88.0 / 92.4 & 92.3 / 85.5
/ 89.8 \\
phenomenon: lists & 284 & 1,217 & 95.7 / 91.2 / 95.7 & 89.9 / 88.2 /
90.9 \\
phenomenon: style & 280 & 955 & 95.5 / 88.5 / 93.1 & 94.1 / 87.2 /
92.1 \\
phenomenon: paraphrase & 245 & 1,150 & 94.9 / 86.9 / 94.1 & 92.9 / 84.6
/ 90.9 \\
phenomenon: identifiers & 239 & 899 & 90.0 / 91.2 / 92.4 & 84.9 / 88.8 /
89.0 \\
phenomenon: noise & 197 & 728 & 96.4 / 87.9 / 92.0 & 95.0 / 87.2 /
90.4 \\
phenomenon: negation & 177 & 850 & 93.0 / 87.9 / 93.7 & 91.1 / 87.0 /
90.6 \\
phenomenon: aliases & 166 & 826 & 96.2 / 89.6 / 97.1 & 92.9 / 88.2 /
93.6 \\
phenomenon: several steps & 142 & 441 & 92.0 / 91.9 / 94.4 & 90.8 / 91.7
/ 93.2 \\
phenomenon: derived values & 84 & 315 & 94.6 / 86.4 / 93.7 & 92.8 / 81.3
/ 86.8 \\
phenomenon: conditional & 17 & 51 & 100.0 / 99.3 / 99.3 & 100.0 / 98.3 /
96.9 \\
\end{longtable}

\noindent\textbf{Extended Data Table 6. Cost per request.} GPU energy of the
whole board. Time, GPU energy and generated tokens per request are means
over all 2,126 requests of the evaluation set. Energy per fully correct request is the GPU energy over the
1,952 complete and out-of-scope requests divided by the number fully
correct. Fully correct: for a complete request, the labelled tool called
with every stated value right and no required argument left empty; for
an out-of-scope request, a decline.

\begin{longtable}[]{@{}
  >{\raggedright\arraybackslash}p{(\columnwidth - 10\tabcolsep) * \real{0.1667}}
  >{\raggedright\arraybackslash}p{(\columnwidth - 10\tabcolsep) * \real{0.1667}}
  >{\raggedright\arraybackslash}p{(\columnwidth - 10\tabcolsep) * \real{0.1667}}
  >{\raggedright\arraybackslash}p{(\columnwidth - 10\tabcolsep) * \real{0.1667}}
  >{\raggedright\arraybackslash}p{(\columnwidth - 10\tabcolsep) * \real{0.1667}}
  >{\raggedright\arraybackslash}p{(\columnwidth - 10\tabcolsep) * \real{0.1667}}@{}}
\toprule\noalign{}
\begin{minipage}[b]{\linewidth}\raggedright
agent
\end{minipage} & \begin{minipage}[b]{\linewidth}\raggedright
s per request (mean)
\end{minipage} & \begin{minipage}[b]{\linewidth}\raggedright
GPU energy per request (J)
\end{minipage} & \begin{minipage}[b]{\linewidth}\raggedright
generated tokens per request
\end{minipage} & \begin{minipage}[b]{\linewidth}\raggedright
GPU energy per fully correct request (Wh)
\end{minipage} & \begin{minipage}[b]{\linewidth}\raggedright
peak GPU memory (MiB)
\end{minipage} \\
\midrule\noalign{}
\endhead
\bottomrule\noalign{}
\endlastfoot
decider-4b & 1.17 & 210 & 13 & 0.072 & 7,368 \\
pplx-decider-flash & 4.59 & 441 & 0 & 0.137 & 12,294 \\
decider-35b-a3b & 2.24 & 413 & 15 & 0.133 & 23,490 \\
OpenJev & 2.56 & 758 & 0 & 0.228 & 21,852 \\
Clef & 3.12 & 927 & 0 & 0.284 & 31,107 \\
pplx-decider-27b & 10.14 & 1,367 & 0 & 0.407 & 35,057 \\
Gemma 4 E4B, reasoning on & 5.85 & 1,710 & 656 & 0.685 & 12,798 \\
Nemotron 3.5 Bolt E4B, reasoning on & 6.15 & 1,801 & 686 & 0.827 &
12,364 \\
Qwen3.5-4B, reasoning on & 3.20 & 933 & 456 & 0.310 & 12,597 \\
Gemma 4 26B-A4B, reasoning on & 7.87 & 2,321 & 711 & 0.753 & 45,341 \\
Qwen3.5-27B, reasoning on & 11.76 & 3,494 & 418 & 1.096 & 36,038 \\
Qwen3.8-27B, reasoning on & 26.09 & 7,793 & 952 & 2.481 & 34,955 \\
Muse Glimmer 30B, reasoning on & 19.82 & 5,924 & 733 & 1.901 & 35,212 \\
Qwen3.5-35B-A3B, reasoning on & 2.60 & 641 & 417 & 0.198 & 27,572 \\
Nemotron 3.5 Bolt 74B-A9B, reasoning on & 5.48 & 1,517 & 584 & 0.467 &
25,515 \\
Gemma 4 E4B, reasoning off & 1.32 & 349 & 132 & 0.135 & 14,208 \\
Nemotron 3.5 Bolt E4B, reasoning off & 1.48 & 398 & 150 & 0.139 &
14,362 \\
Qwen3.5-4B, reasoning off & 1.22 & 337 & 164 & 0.117 & 12,044 \\
Gemma 4 26B-A4B, reasoning off & 1.49 & 405 & 123 & 0.142 & 45,532 \\
Qwen3.5-27B, reasoning off & 4.54 & 1,330 & 156 & 0.413 & 36,002 \\
Qwen3.8-27B, reasoning off & 3.71 & 1,081 & 127 & 0.363 & 35,976 \\
Muse Glimmer 30B, reasoning off & 9.25 & 2,751 & 340 & 0.853 & 36,030 \\
Qwen3.5-35B-A3B, reasoning off & 1.07 & 247 & 157 & 0.080 & 27,573 \\
Nemotron 3.5 Bolt 74B-A9B, reasoning off & 1.87 & 461 & 169 & 0.166 &
26,930 \\
\end{longtable}

\noindent\textbf{Extended Data Table 7. Fabricated content per agent.} Adjusted:
flagged calls corrected by the precision of the independent review
(lower-upper bound).

\begin{longtable}[]{@{}
  >{\raggedright\arraybackslash}p{(\columnwidth - 10\tabcolsep) * \real{0.1667}}
  >{\raggedright\arraybackslash}p{(\columnwidth - 10\tabcolsep) * \real{0.1667}}
  >{\raggedright\arraybackslash}p{(\columnwidth - 10\tabcolsep) * \real{0.1667}}
  >{\raggedright\arraybackslash}p{(\columnwidth - 10\tabcolsep) * \real{0.1667}}
  >{\raggedright\arraybackslash}p{(\columnwidth - 10\tabcolsep) * \real{0.1667}}
  >{\raggedright\arraybackslash}p{(\columnwidth - 10\tabcolsep) * \real{0.1667}}@{}}
\toprule\noalign{}
\begin{minipage}[b]{\linewidth}\raggedright
agent
\end{minipage} & \begin{minipage}[b]{\linewidth}\raggedright
calls
\end{minipage} & \begin{minipage}[b]{\linewidth}\raggedright
nonexistent argument (calls)
\end{minipage} & \begin{minipage}[b]{\linewidth}\raggedright
calls with an invented value, adjusted (\%)
\end{minipage} & \begin{minipage}[b]{\linewidth}\raggedright
underspecified requests: required value filled with no source / with a
value written elsewhere
\end{minipage} & \begin{minipage}[b]{\linewidth}\raggedright
schema-invalid calls
\end{minipage} \\
\midrule\noalign{}
\endhead
\bottomrule\noalign{}
\endlastfoot
decider-4b & 1,807 & 0 & 0 & 0 / 5 & 7 \\
pplx-decider-flash & 1,832 & 0 & 0 & 0 / 6 & 5 \\
decider-35b-a3b & 1,796 & 0 & 0 & 0 / 2 & 3 \\
OpenJev & 1,814 & 0 & 0 & 0 / 3 & 1 \\
Clef & 1,811 & 0 & 0 & 0 / 3 & 6 \\
pplx-decider-27b & 1,828 & 0 & 0 & 0 / 3 & 0 \\
Gemma 4 E4B, reasoning on & 1,266 & 23 & 5.5--5.7 & 0 / 1 & 1 \\
Nemotron 3.5 Bolt E4B, reasoning on & 1,061 & 19 & 8.5--8.8 & 0 / 0 &
4 \\
Qwen3.5-4B, reasoning on & 1,747 & 0 & 17.3--17.9 & 28 / 7 & 4 \\
Gemma 4 26B-A4B, reasoning on & 1,732 & 22 & 11.0--11.4 & 16 / 7 & 4 \\
Qwen3.5-27B, reasoning on & 1,751 & 0 & 10.0--10.3 & 31 / 7 & 2 \\
Qwen3.8-27B, reasoning on & 1,832 & 0 & 13.0--13.4 & 54 / 10 & 0 \\
Muse Glimmer 30B, reasoning on & 1,803 & 0 & 18.6--19.2 & 33 / 15 & 7 \\
Qwen3.5-35B-A3B, reasoning on & 1,702 & 0 & 9.6--10.0 & 19 / 7 & 3 \\
Nemotron 3.5 Bolt 74B-A9B, reasoning on & 1,728 & 0 & 13.0--13.4 & 16 /
8 & 1 \\
Gemma 4 E4B, reasoning off & 1,465 & 53 & 13.2--18.6 & 12 / 8 & 9 \\
Nemotron 3.5 Bolt E4B, reasoning off & 1,691 & 51 & 19.7--27.8 & 34 / 17
& 14 \\
Qwen3.5-4B, reasoning off & 1,828 & 0 & 16.8--23.7 & 97 / 14 & 7 \\
Gemma 4 26B-A4B, reasoning off & 1,805 & 33 & 13.3--18.8 & 99 / 14 &
15 \\
Qwen3.5-27B, reasoning off & 1,761 & 0 & 7.4--10.4 & 48 / 15 & 2 \\
Qwen3.8-27B, reasoning off & 1,982 & 0 & 9.6--13.5 & 108 / 22 & 2 \\
Muse Glimmer 30B, reasoning off & 1,745 & 0 & 10.4--14.7 & 19 / 14 &
7 \\
Qwen3.5-35B-A3B, reasoning off & 1,817 & 0 & 9.9--13.9 & 85 / 21 & 1 \\
Nemotron 3.5 Bolt 74B-A9B, reasoning off & 1,927 & 0 & 15.5--21.8 & 91 /
47 & 3 \\
\end{longtable}

\textbf{Extended Data Table 8. Pairwise comparisons by reasoning mode
and under scoring variants.} Pairs with System 1 significantly ahead /
LLM significantly ahead / no significant difference (Holm-adjusted P
\textless{} 0.05; one correction over the 108 pairs of each measure,
whose 54 pairs with reasoning on and 54 with reasoning off are shown in
separate rows; the better-mode row has its own correction over its 54
pairs). The
better mode of each LLM is chosen on its fully correct requests. Scoring
variants (Methods): search text compared as written instead of after
removing stop words; model names of the PyCaret regression server
matched through PyCaret's name table instead of as written. Fully
correct: for a complete request, the labelled tool called with every
stated value right and no required argument left empty; for an
out-of-scope request, a decline.

\begin{longtable}[]{@{}
  >{\raggedright\arraybackslash}p{(\columnwidth - 6\tabcolsep) * \real{0.2500}}
  >{\raggedright\arraybackslash}p{(\columnwidth - 6\tabcolsep) * \real{0.2500}}
  >{\raggedright\arraybackslash}p{(\columnwidth - 6\tabcolsep) * \real{0.2500}}
  >{\raggedright\arraybackslash}p{(\columnwidth - 6\tabcolsep) * \real{0.2500}}@{}}
\toprule\noalign{}
\begin{minipage}[b]{\linewidth}\raggedright
analysis
\end{minipage} & \begin{minipage}[b]{\linewidth}\raggedright
fully correct
\end{minipage} & \begin{minipage}[b]{\linewidth}\raggedright
tool calling
\end{minipage} & \begin{minipage}[b]{\linewidth}\raggedright
argument filling
\end{minipage} \\
\midrule\noalign{}
\endhead
\bottomrule\noalign{}
\endlastfoot
main, all pairs & 73 / 11 / 24 & 76 / 5 / 27 & 46 / 19 / 43 \\
main, reasoning on & 37 / 6 / 11 & 41 / 2 / 11 & 23 / 8 / 23 \\
main, reasoning off & 36 / 5 / 13 & 35 / 3 / 16 & 23 / 11 / 20 \\
each LLM in its better mode (54 pairs) & 33 / 8 / 13 & 33 / 4 / 17 & 19
/ 12 / 23 \\
scoring variant: search text compared as written & 73 / 11 / 24 & 76 / 5
/ 27 & 46 / 19 / 43 \\
scoring variant: AutoML model names through PyCaret's name table & 73 /
11 / 24 & 76 / 5 / 27 & 45 / 19 / 44 \\
\end{longtable}

\endgroup
\end{document}